\documentclass[letterpaper, 10 pt, conference]{ieeeconf}

\IEEEoverridecommandlockouts                              %

\usepackage{graphicx} %
\usepackage{balance}
\usepackage{algorithm,algorithmic}

\title{\LARGE \bf
Fast Local Planning and Mapping in Unknown Off-Road Terrain
}

\author{Timothy Overbye and Srikanth Saripalli%
\thanks{$^{1}$Tim Overbye and Srikanth Saripalli are with the Department of Mechanical Engineering, Texas A\&M University, College Station, TX 77840, USA
        {\tt\small overbye2@tamu.edu}
        {\tt\small ssaripalli@tamu.edu}}%
}

\begin{document}

\maketitle
\thispagestyle{empty}
\pagestyle{empty}

\begin{abstract}

In this paper, we present a fast, on-line mapping and planning solution for operation in unknown, off-road, environments. We combine obstacle detection along with a terrain gradient map to make simple and adaptable cost map. This map can be created and updated at 10~Hz. An A* planner finds optimal paths over the map. Finally, we take multiple samples over the control input space and do a kinematic forward simulation to generated feasible trajectories. Then the most optimal trajectory, as determined by the cost map and proximity to A* path, is chosen and sent to the controller. Our method allows real time operation at rates of 30~Hz. We demonstrate the efficiency of our method in various off-road terrain at high speed.

\end{abstract}

\section{INTRODUCTION}

\begin{figure}[htpb]
      \centering
      \framebox{\parbox{3in}{
      
      \includegraphics[width=\linewidth]{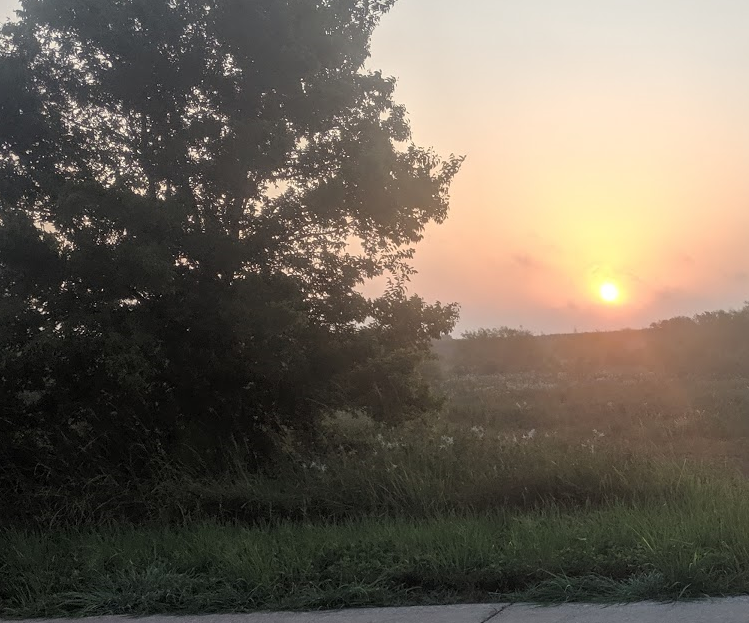}
	}}
      
      \caption{An example of off-road terrain at the Texas A\&M Rellis Campus.}
      \label{offroad_terrain}
   \end{figure}

Navigating off-road environments is a unique challenge for unmanned ground vehicles. This is, in part, due to the unstructured nature of an off-road environment. Terrain that can't be driven through, such as swamps and deep mud, may initially appear passable. Conversely, objects like tall grass and small bushes (as in figure~\ref{offroad_terrain}) will appear as obstacles even if they can be driven through.
Additionally, in smooth roads, and similar environments, obstacles can be relatively easily detected and avoided. This is not always the case in rough terrain where we may be able to drive over perceived obstacles, such as small rocks and fallen trees, rather than around. Our strategy to address this is to not only track obstacles but also measure local terrain gradients. These maps can then be combined into a composite cost map as done by~\cite{terrain_maps}. This method allows us to drive over some obstacles if no other path is available, although it does not solve the mud or swamp cases. 

We then present a local planner that is able to make kinematically feasible paths across the composite map. These paths need not be optimal but rather collision free and achievable by the vehicle yet computationally fast such that that they can be computed in real time. The goal of this paper is to present a real-time, off-road, local mapping and planning algorithm capable of operation in an unknown environment. We desire our vehicle to perform at its limits. Our test vehicle has a top speed of 4.5 m/s and we ran tests from 3-4 m/s.

   \begin{figure}[htpb]
      \centering
      \framebox{\parbox{3in}{
      
      \includegraphics[width=\linewidth]{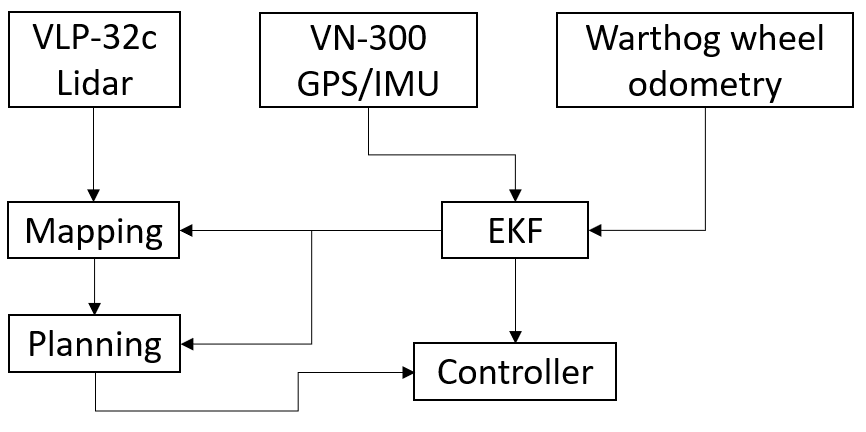}
	}}
      
      \caption{System diagram of the Warthog. An extended Kalman filter fuses odometry data from wheel sensors with data from the GPS/IMU. The odometry, and a point cloud from lidar, are used to make a local map. Finally, the planner uses the map to create a path which is sent to the vehicle controller. }
      \label{system_diagram}
   \end{figure}

   \begin{figure}[htpb]
      \centering
      \framebox{\parbox{3in}{
      
      \includegraphics[width=\linewidth]{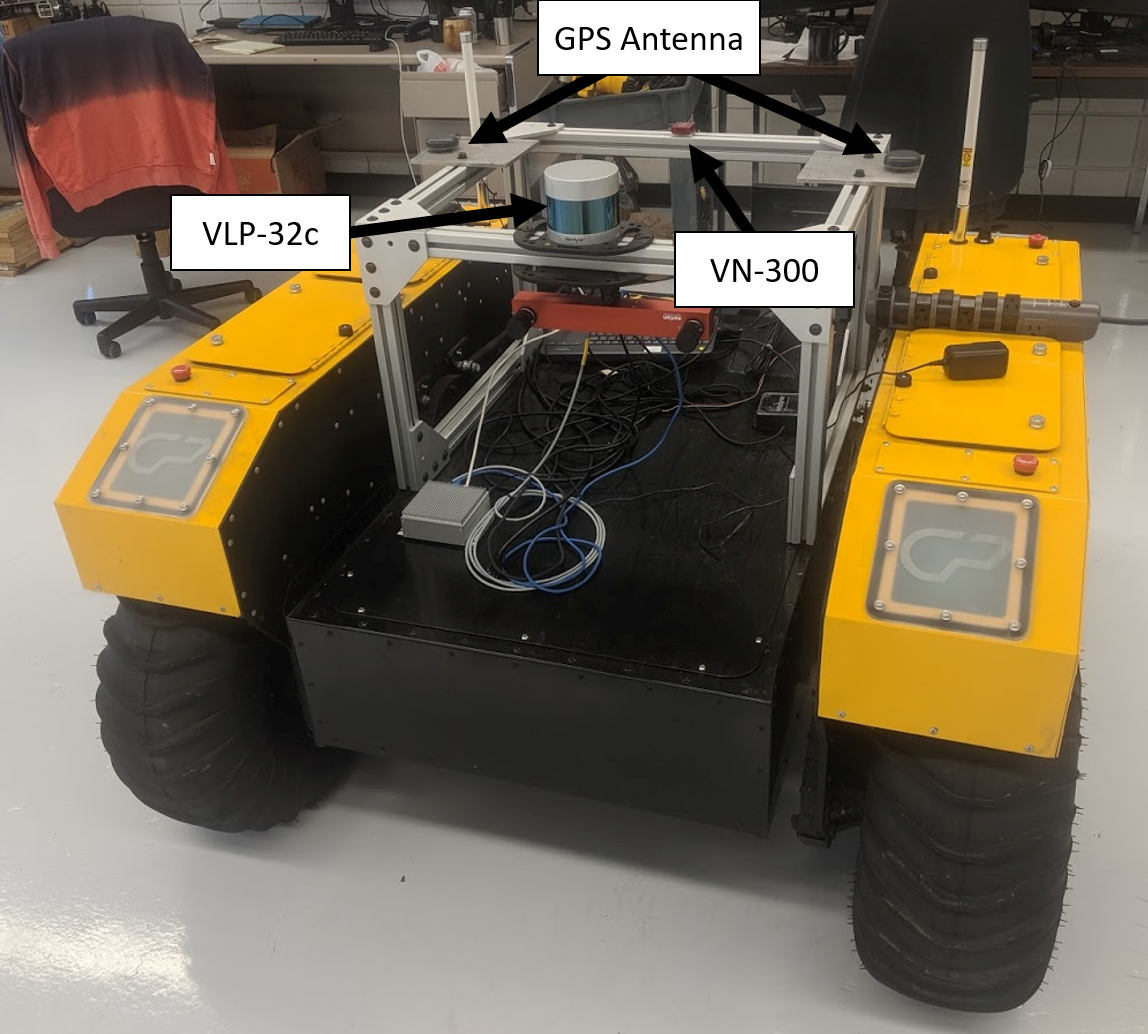}
	}}
      
      \caption{The Warthog showing mounting of the VLP-32c lidar and VN-300 GPS/IMU.}
      \label{warthog_sensors}
   \end{figure}
   
\section{SYSTEM DESCRIPTION}
Figure~\ref{system_diagram} shows a system overview of our test platform. The vehicle is a Clearpath Robotics Warthog with a top speed of 4.5 m/s. We mounted a VLP-32c lidar and VN-300 GPS/IMU as show in figure~\ref{warthog_sensors}. The wheel odometry and GPS/IMU date are fused by an extended Kalman filter to provide an accurate position estimate. This position, along with data from the VLP-32c lidar is used to create the local map. Then a plan is made over this map and is sent to the vehicle controller.

\section{RELATED WORK}
Much work has already been done on the local planning problem~\cite{planning_overview}\cite{yang2016survey}\cite{rrt}\cite{rrt_star}\cite{a_star}\cite{star_overview}\cite{potential_field}\cite{potential_field2}\cite{dwa}\cite{dwa2}. Most approaches can be broken down into four main categories, sample-based, graph-based, potential-field approaches, and variations on the dynamic window approach. 

Sample-based planners, such as rapidly-exploring random tree (RRT)~\cite{rrt} and its derivatives~\cite{rrt_star}, can explore the space very efficiently. They are also well suited to high dimensional spaces and nonholonomic constraints. As such, they are capable of satisfying the kinematic (and sometimes dynamic) constraints~\cite{rrt_dynamics} of ground vehicles. However, even in the best case there is only a guarantee of optimality in the asymptotic case~\cite{rrt_optmality}. Additionally, there is an explicit trade-off between processing time and path quality. 

Graph-based planners guarantee an optimal solution over the graph~\cite{a_star}\cite{star_overview}. To build the graph they require a discretized representation of the vehicle’s state space. For holonomic constraints this works well and can be easily solved. However, nonholonomic constraints cause the graph size to quickly reach unsolvable sizes.

Potential-field planners are simple and easy to implement~\cite{potential_field}\cite{potential_field2}. Although not guaranteed optimality  they can very quickly find good enough paths. The potential-field planner creates a vector field pointing towards the goal. Then obstacles are given a repulsive field and gradient decent is done over the combined field. However, cost maps that are continuous do not have an easily defined repulsive direction and are thus difficult to integrate.

The dynamic window approach samples from a family of trajectories over a short time window~\cite{dwa}\cite{dwa2}. This guarantees kinematically feasible and optimal paths within the time window. But these paths are limited to the length of the time window~\cite{dwa}. However, by only sampling within time window a far greater sample density is allowed. This higher density ensures that the optimal path will likely be sampled. However, if the time window is increased this path density drops off. This leads to a trade off between good paths and long paths. Additionally, as with potential-field planners, it is difficult to integrate a non-binary cost map with this approach as it evaluates paths by distance and collision.

Our method uses the A* planner~\cite{a_star} as a base due to its fast processing time and optimality. Other graph-based planners, such as D* or D* Lite~\cite{star_overview} could have been used for this step. However, given the processing speed of A* over our map (30~Hz), the efficient replanning of D* was unnecessary. Next, we sample the control input space of the vehicle over a time horizon to generate trajectories that follow the A* path. Both steps are done over an easily modifiable cost map incorporating obstacles, movement costs, and any other custom cost functions. This approach is fast to compute such that the entire path can be replanned in response to new information in real time. Due to the trajectory planner, the path given to the controller is guaranteed to be kinematically feasible. However, this means that we loose the optimality of the A* path. Additionally, the trajectory planner is a greedy planner with respect to its endpoint. Thus, when the endpoint cannot follow the A* path due to various constrains it can potentially get stuck. Nevertheless, in practice the resulting trajectory follows the A* path in almost all cases.

\section{MAPPING}
\begin{figure}[htpb]
      \centering
      \framebox{\parbox{3in}{
      
      \includegraphics[width=\linewidth]{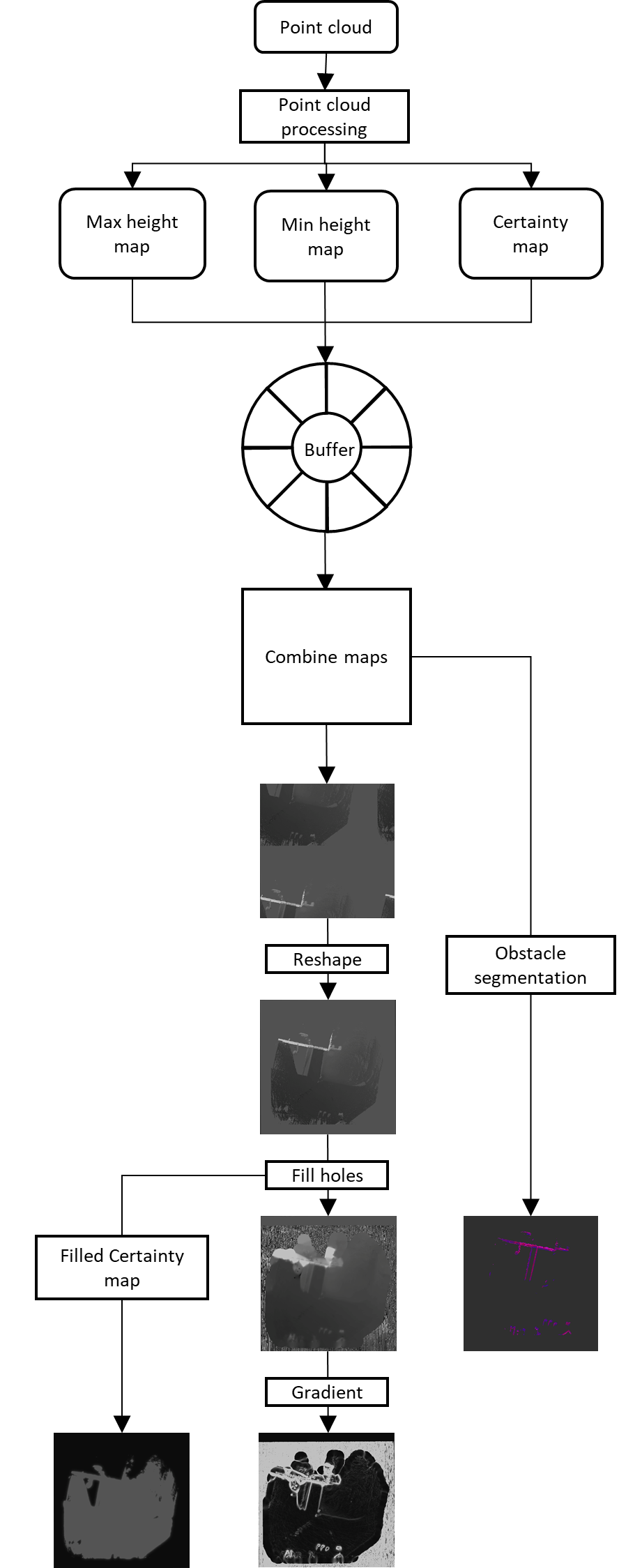}
	}}
      
      \caption{Flowchart showing the mapping process. The point cloud is processed into maps that are inserted into a buffer. During map processing maps in this buffer are combined and then processed into the final output maps.}
      \label{mapping_flowchart}
   \end{figure}

The mapping module creates a representation of the local terrain. However, due to the limited resolution of even the best sensors data must be accumulated over multiple scans. This necessitates some method of storing and processing the accumulated data. Figure~\ref{mapping_flowchart} shows an overview of this process.

\subsection{Map Storage}
\begin{figure}[htpb]
      \centering
      \framebox{\parbox{3in}{
      
      \includegraphics[width=\linewidth]{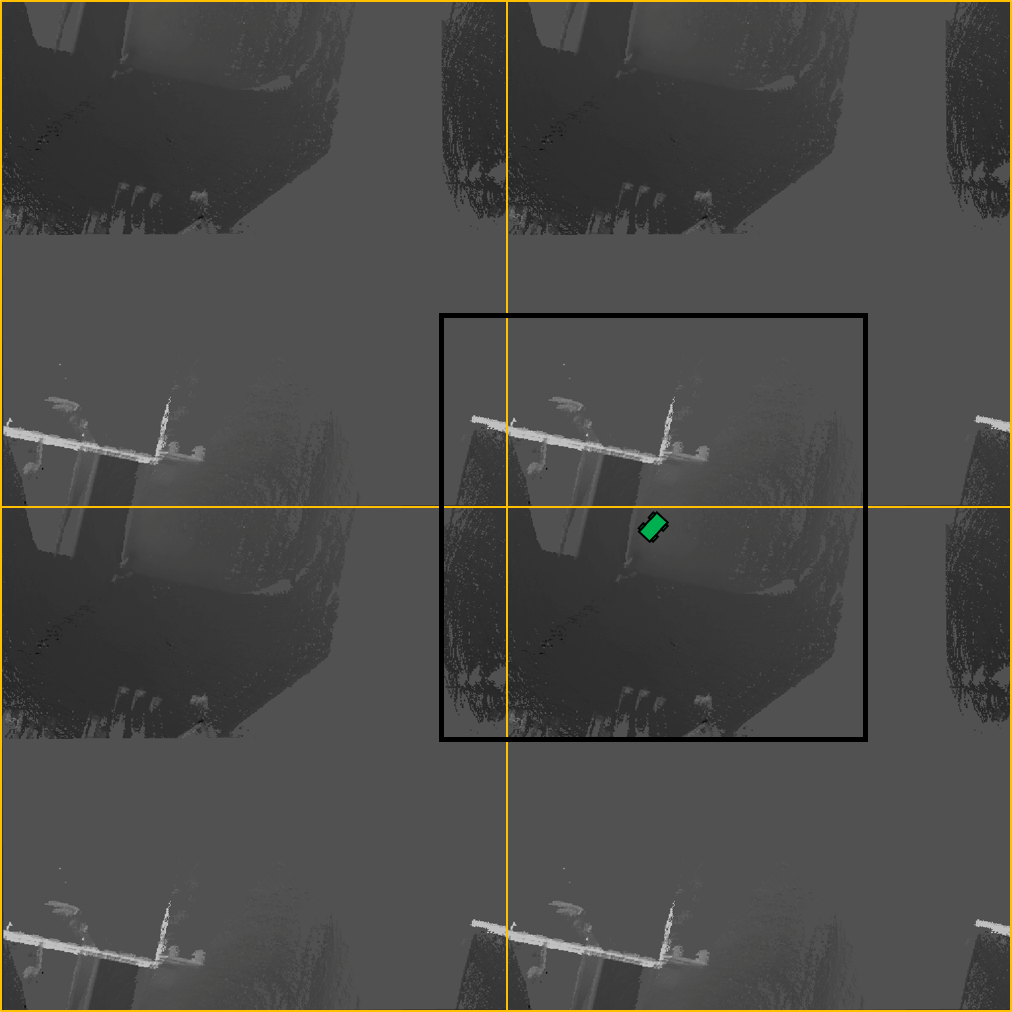}
	}}
      
      \caption{The wrapping map is static while the vehicle (green) moves over it. Any point outside the map is wrapped back into it. This is equivalent to surrounding the map with copies of itself as shown here. Data is only contained within the ROI (black square). As the vehicle moves data outside the ROI is cleared.}
      \label{wrap_fig}
   \end{figure}

A square grid-based map is defined by a length and resolution (we used 512x512 grid with a 0.2~m resolution to give a radius of about 50~m). This map is stored as a wrappable map where the map is stationary and the vehicle moves across it. The vehicle pose and sensor data can be mapped to the same memory using only 2d modulo operations. However, due to the cyclic nature of the map there is no one-to-one mapping from map space to world space. This necessitates the use of a fixed size region of interest (ROI) around the vehicle, shown in figure~\ref{wrap_fig}. All data within the ROI is assumed to be valid and all data outside the ROI is cleared during map update. This ROI must be smaller than the total map size and is given by $v_{max} f$ where $v_{max}$ is the maximum vehicle speed and $f$ is the map update frequency. This ensures that no invalid data will "jump" through the region outside the ROI when the vehicle moves. The primary benefit of this method is that, if the ROI update is applied to all maps in memory, they can be combined through a simple per cell weighted average as all cells with valid data correspond to the same world space.

To store data from multiple scans several of these maps are stored in a circular buffer. This allows more recent data to be weighted more than older data.

\subsection{Point Cloud Processing}

During the point cloud processing four temporary maps of the same size and shape as the main map are created. A maximum and minimum height map,a certainty map, and an obstacle map. The maximum and minimum height maps are initialized to negative infinity and positive infinity respectively. The certainty map and obstacle map are initialized to zero.

\begin{algorithm}
 \caption{Algorithm for map creation from point cloud}
 \begin{algorithmic}[1]
 \label{alg:map_creation}
 \renewcommand{\algorithmicrequire}{\textbf{Input:}}
 \renewcommand{\algorithmicensure}{\textbf{Output:}}
  \FOR {$point$ in $pointcloud$}
  \IF {$distance(point,robot) < ROI\_radius$}
  \STATE $x = int( point.x / map\_resolution ) \% map\_size$
  \STATE $y = int( point.y / map\_resolution ) \% map\_size$

  \IF {$max\_height[x,y] < point.z$}  
  \STATE $max\_height[x,y] = point.z$
  \STATE $certainty[x,y] = 1$
  \ENDIF
  
  \IF {$min\_height[x,y] > point.z$}  
  \STATE $min\_height[x,y] = point.z$
  \STATE $certainty[x,y] = 1$
  \ENDIF
   
  \ENDIF
  \ENDFOR 
 \end{algorithmic} 
 \end{algorithm}

Algorithm~\ref{alg:map_creation} describes the creation of the height maps and certainty map.

After all the points have been processed the obstacle map is created. For each cell that data was collected in the difference between the maximum and minimum height is compared to a threshold (we used 0.5~m). If the difference is greater than this threshold the cell is marked as an obstacle.

Finally, all the temporary maps are added into main buffer at the current buffer index. This process is completed at the sensor's sample rate, 10~hertz in our case.

\subsection{Map Processing}
In the map processing phase first the maps in the buffer are combined using a weighted average of each point. This is described by equations~\ref{equ:map_making1} and~\ref{equ:map_making2}.

\begin{equation}
\label{equ:map_making1}
H[x,y] =  \frac{\sum_{i=0} ^n h_i[x,y] c_i[x,y] f(i)}{\sum_{i=0} ^n c_i[x,y] f(i)}
\end{equation}
\begin{equation}
\label{equ:map_making2}
f(i) = \frac{n-i}{n}
\end{equation}

\begin{figure}[htpb]
      \centering
      \framebox{\parbox{3in}{
      
      \includegraphics[width=\linewidth]{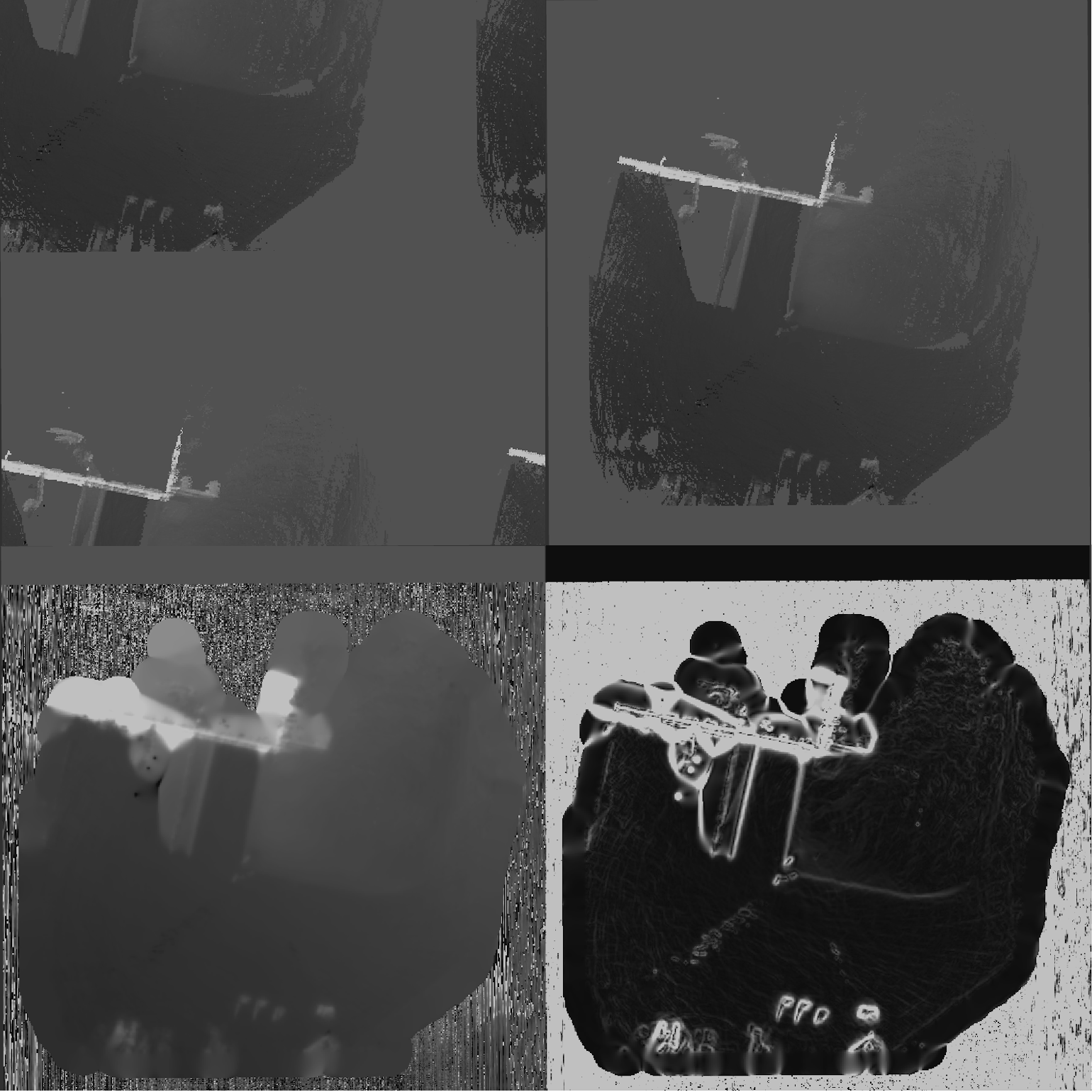}
      \includegraphics[width=\linewidth]{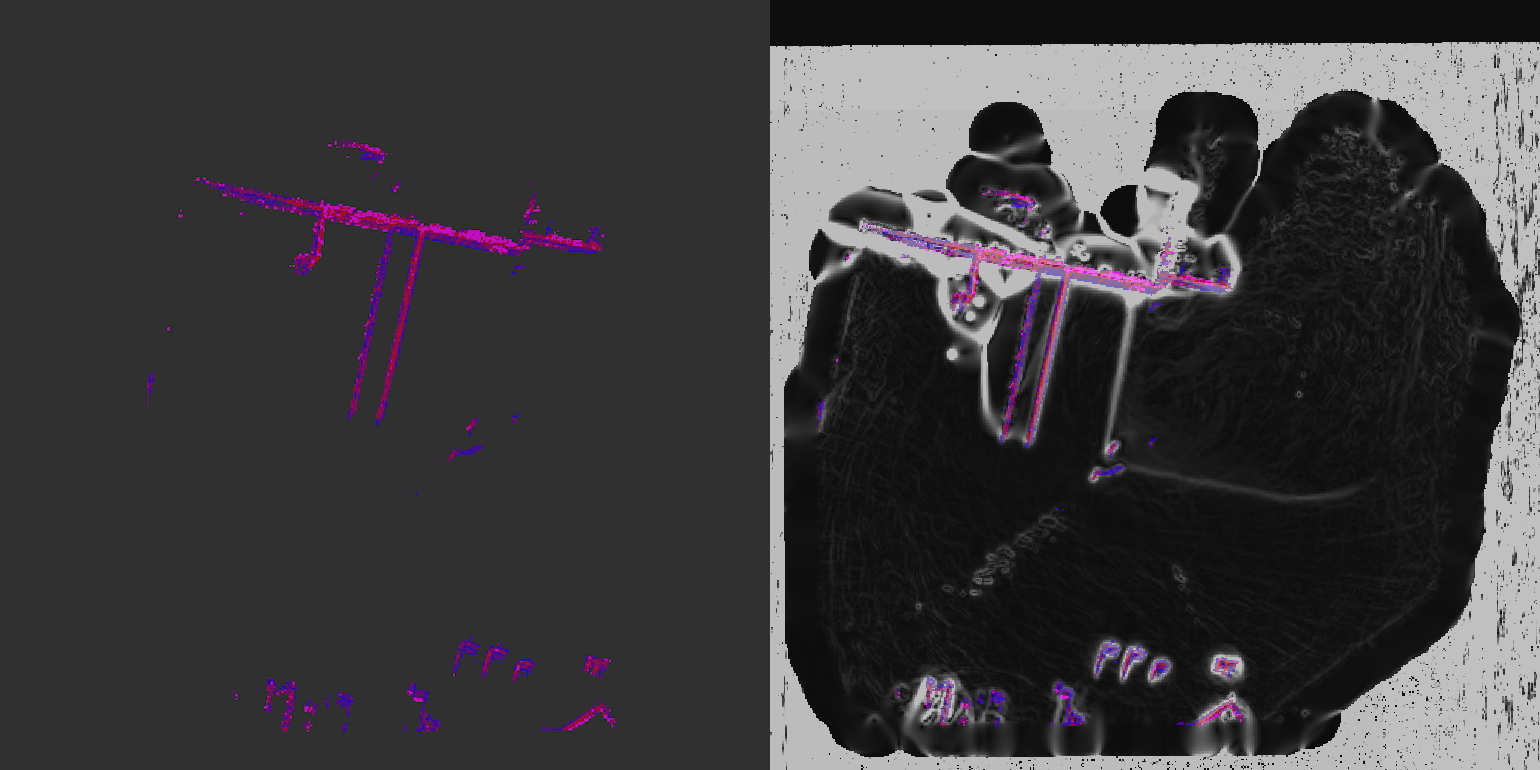}
	}}
      
      \caption{Maps at the main stages of the map processing process. From left to right, top down: wrapped height map, vehicle centered height map, height map with holes filled, gradient map, obstacle map, and combined obstacle and gradient map. }
      \label{process_fig}
   \end{figure}

Where $H$ is the outputted height map, $h_i$ is the buffered height map at index $i$, $c_i$ is the certainty map at index $i$, $n$ is the buffer size, and $i=0$ is the most recent buffer index. Figure~\ref{process_fig} (top left) shows the output of this process.

Next, the map is reshaped such that it is centered on the vehicle (figure~\ref{process_fig} top right).

At this point the obstacle map is completed and published (figure ~\ref{process_fig} bottom left). A normalized convolution\cite{normal_convolution} is used to fill holes in both the height map and certainty map (figure ~\ref{process_fig} middle left). Finally the numerical gradient of the filled height map is taken and published (figure~\ref{process_fig} middle right). Based on~\cite{terrain_maps}.

   \begin{figure}[htpb]
      \centering
      \framebox{\parbox{3in}{
      
      \includegraphics[width=\linewidth]{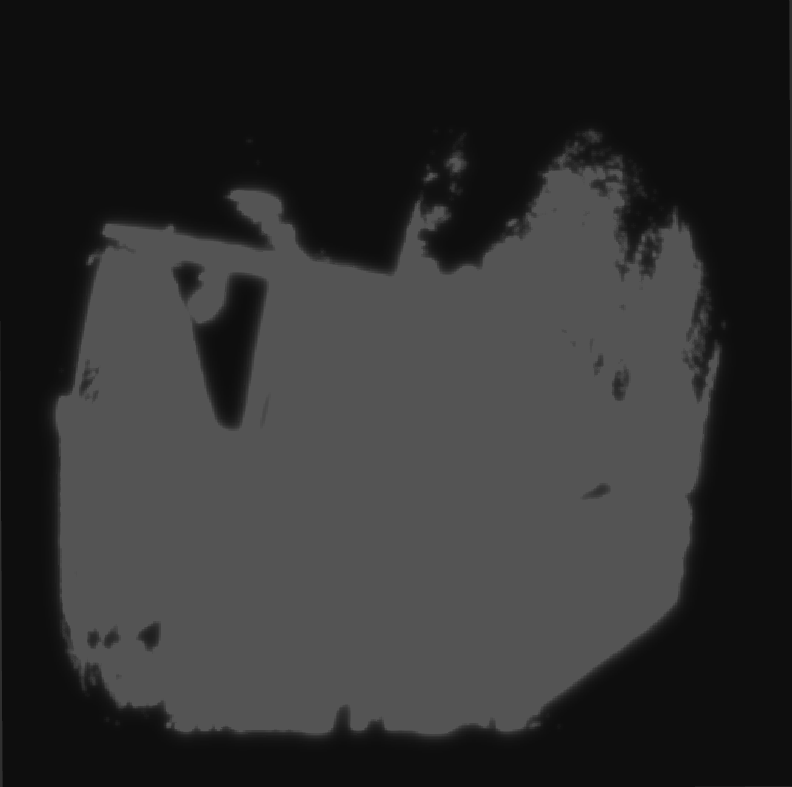}
	}}
      
      \caption{The outputted certainty map after hole filling.}
      \label{cert_fig}
   \end{figure}

\section{PLANNING}
This system uses an implementation of A* for medium range planning (to the edge of the local map) and a kinematic based trajectory sampling approach for local control. A* ensures that the robot does not get stuck in local minima within the map and the trajectory sampler ensures kinetically feasible paths. 

\subsection{Cost Map Creation}

Before any planning can be done the local cost map must be created. This is a function of the obstacle map, certainty map, and gradient map. Finally, the map is expanded by the radius of the robot.

\subsection{A*}
A* is the first layer of the planning algorithm. We chose it due to its guarantee of optimality and fast run time. In our implementation we used 8 way connectivity, the euclidean distance as a heuristic, and an update rate of 5~Hz. The robot’s position is used as the start of the search. If the goal point is within the local map it is used as the goal for A*. Otherwise, the closest point within the map to the goal is used for A*.

\subsection{Trajectory Sampling}
\begin{figure}[htpb]
      \centering
      \framebox{\parbox{3in}{
      
      \includegraphics[width=\linewidth]{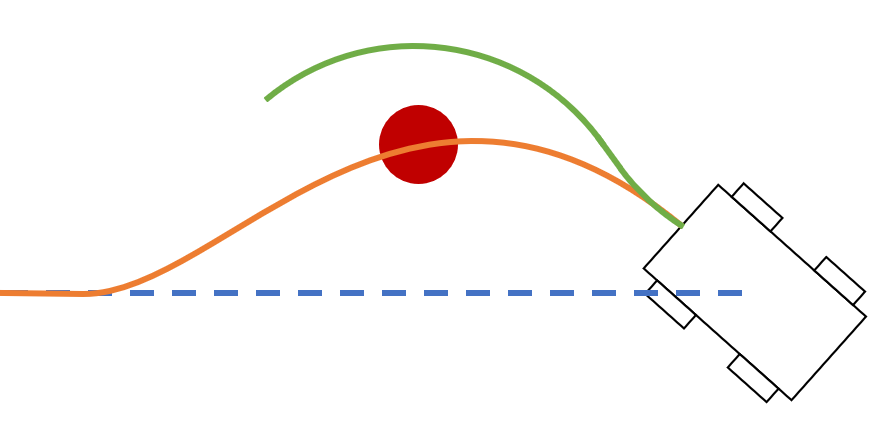}
	}}
      
      \caption{Example of pure pursuit (orange) and trajectory sampling (green) following a path (blue dashed) with an obstacle (red). Pure pursuit collides with the obstacle while trajectory sampling does not.}
      \label{ppvts_fig}
   \end{figure}

Although the A* path is guaranteed to be optimal it is most likely not achievable due to constraints on the robot’s motion. For example, if the goal point is directly to the left of the robot the A* path will be a straight line. However, following this path requires a robot with a turning radius of zero. A simple approach to this problem is the pure pursuit controller. However, pure pursuit will often cut corners and has no information about the cost map. Due to this, in obstacle dense environments, such as a forest, pure pursuit of the A* path will lead to collisions. Figure~\ref{ppvts_fig} shows an example where a collision free, but unachievable path is followed by pure pursuit and our trajectory sampling. Pure pursuit, having no knowledge of obstacles, collides with the red obstacle. Instead, our approach is to generate many kinematically feasible paths then select the most optimal path. 

\begin{figure}[htpb]
      \centering
      \framebox{\parbox{3in}{
      
      \includegraphics[width=\linewidth]{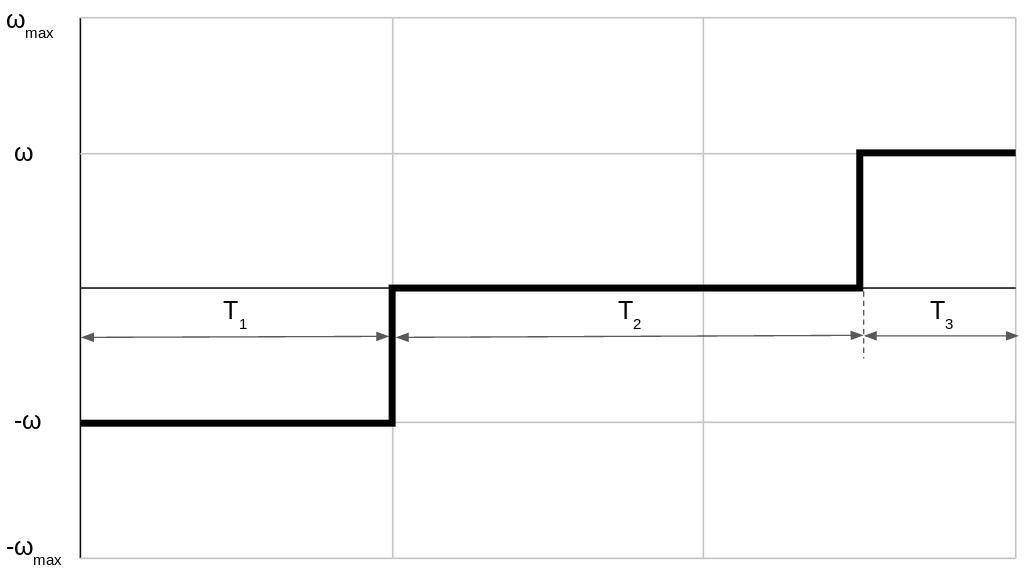}
	}}
      
      \caption{An example of a sample over the controller space. The y axis is yaw rate and the x axis is time. This shows a trajectory where the vehicle turns right at rate $\omega$ for time $T_1$, goes straight for time $T_2$, then turns left at rate $\omega$ for time $T_3$.}
      \label{controll_fig}
   \end{figure}

\begin{figure}[htpb]
      \centering
      \framebox{\parbox{3in}{
      
      \includegraphics[width=\linewidth]{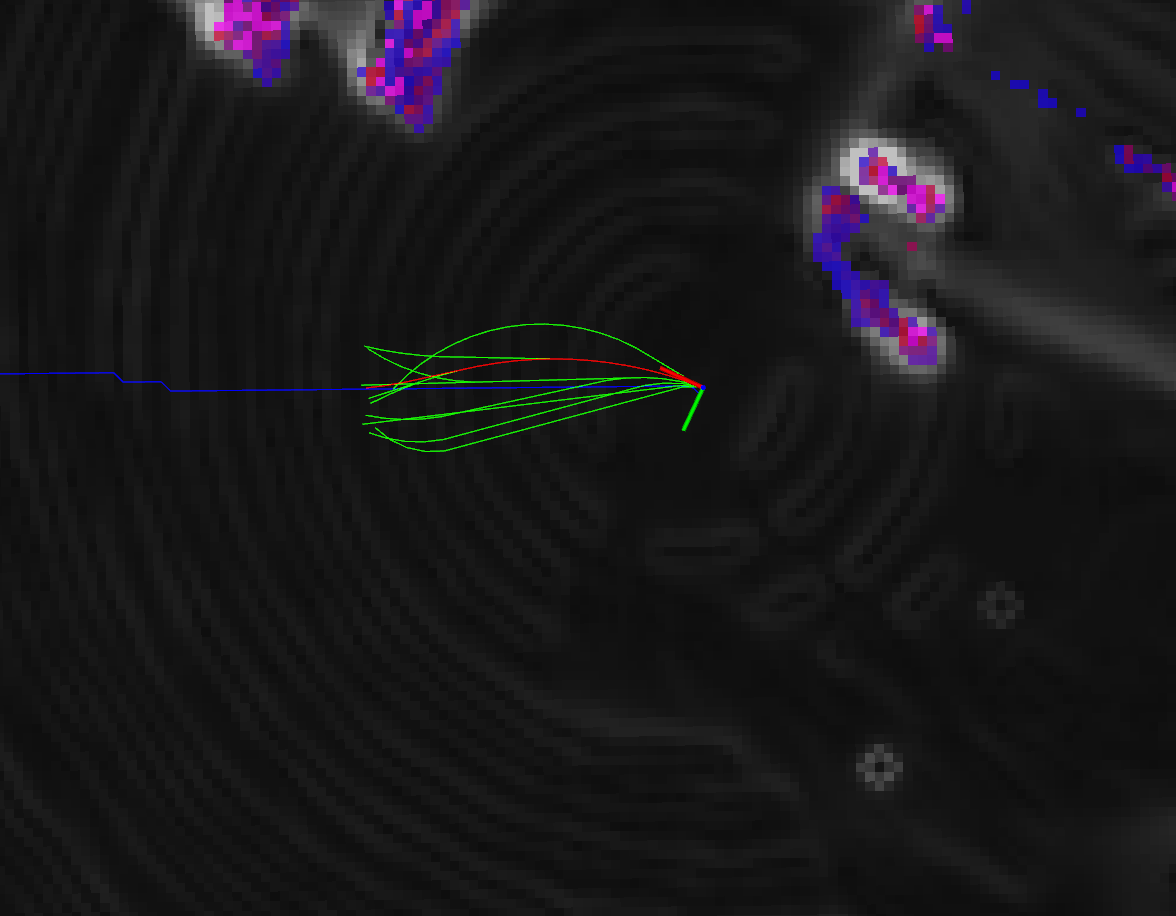}
	}}
      
      \caption{The selected trajectory (red) and 10 low cost candidates (green) following the A* path (blue).}
      \label{paths_fig}
   \end{figure}

We generate the paths by sampling the space of possible control inputs over a short time horizon $T$ (3 seconds in our case). To reduce the sample space we assume the robot is moving at a constant speed and sample over possible angular velocities. All sampled trajectories follow the same structure. First the robot turns right with a given angular velocity $\omega$ for some duration $T_1$. Next, the robot goes straight for duration $T_2$. Finally the robot turns left with velocity $\omega$ for duration $T_3$. The same procedure is done again for the left-straight-right case. $\omega$ is calculated by linearly sampling angular velocities from zero to the robot's maximum turn rate $\omega_{max}$. $T_1$ and $T_2$ are linearly sampled over the time horizon with $T_3$ selected such that $T_1 + T_2 + T_3 = T$. This gives a 3d space to sample over $\omega$, $T_1$, $T_2$. We chose 10 samples over each dimension giving about 1000 possible trajectories. Figure~\ref{controll_fig} shows an example trajectory in the controller space. Figure~\ref{paths_fig} shows the selected trajectory along with a subset of the other candidates.

We select the target point for trajectory evaluation by pure pursuit along the A* path with a lookahead distance of $T V_{max}$ where $V_{max}$ is the robot's maximum speed. Each trajectory $t$ is evaluated by the following cost function described in equation~\ref{equ:trajectory_cost}.
\begin{equation}
\label{equ:trajectory_cost}
C(t) = Map(t) + \alpha \frac{\omega}{\omega_{max}} + \beta Dist(t) + \gamma H_{err}(t)
\end{equation}
Where $Map(t)$ is the sum of the cost of all the grid cells passed through by $t$. $Dist(t)$ is the distance between the end of t and the goal point. $H_{err}(t)$ is the heading error between the last point of $t$ and the goal point. $\frac{\omega}{\omega_{max}}$  represents the curvature of the path with a penalty for higher curvatures. And $\alpha$, $\beta$, and $\gamma$ are weighting factors. Note that this cost function can be quickly computed. This allows us to compute and evaluate trajectories at a fast rate (30 hertz in our case).

\section{RESULTS}

   \begin{figure}[t]
      \centering
      \framebox{\parbox{3in}{
      
      \includegraphics[width=\linewidth]{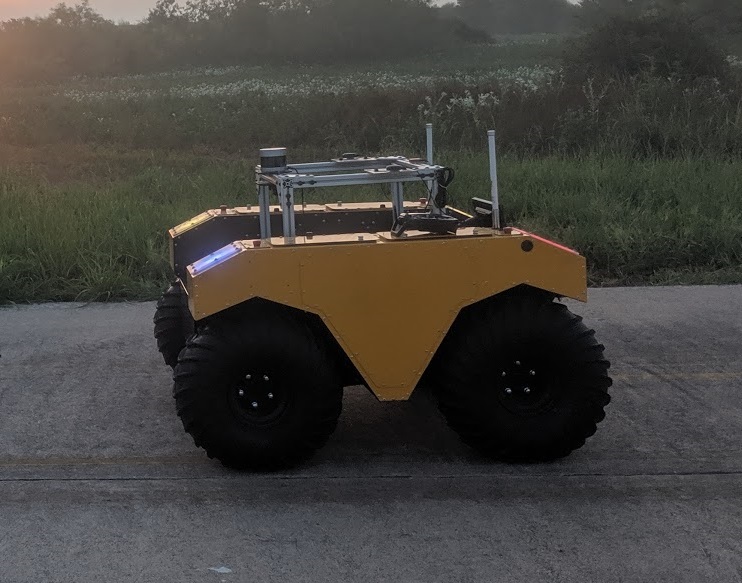}
	}}
      
      \caption{The Warthog is a differential drive off-road robot shown here with lidar and GPS/IMU mounted}
      \label{warthog}
   \end{figure}
   
   \begin{figure}[h!]
      \centering
      \framebox{\parbox{3in}{
      
      \includegraphics[width=\linewidth]{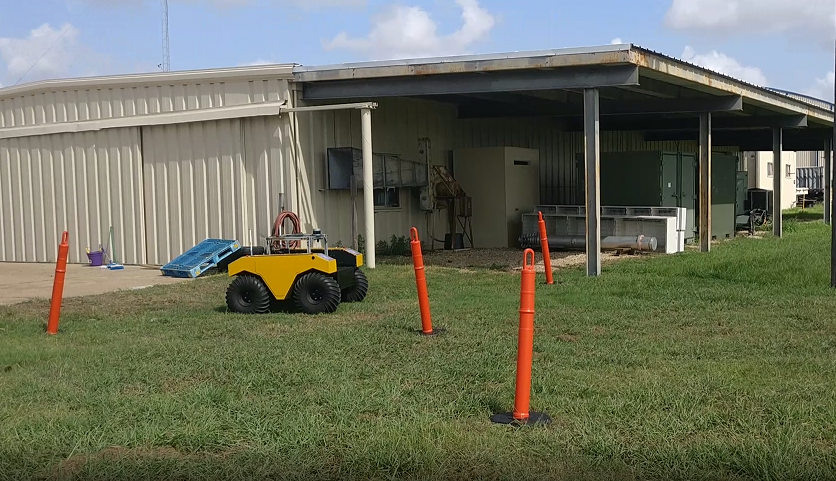}
	}}
      
      \caption{Texas A\&M Rellis Campus test area used in figure~\ref{test_1}.}
      \label{rellis}
   \end{figure}   
   
  	\begin{figure}[thpb]
      \centering
      \framebox{\parbox{3in}{
      \includegraphics[width=\linewidth]{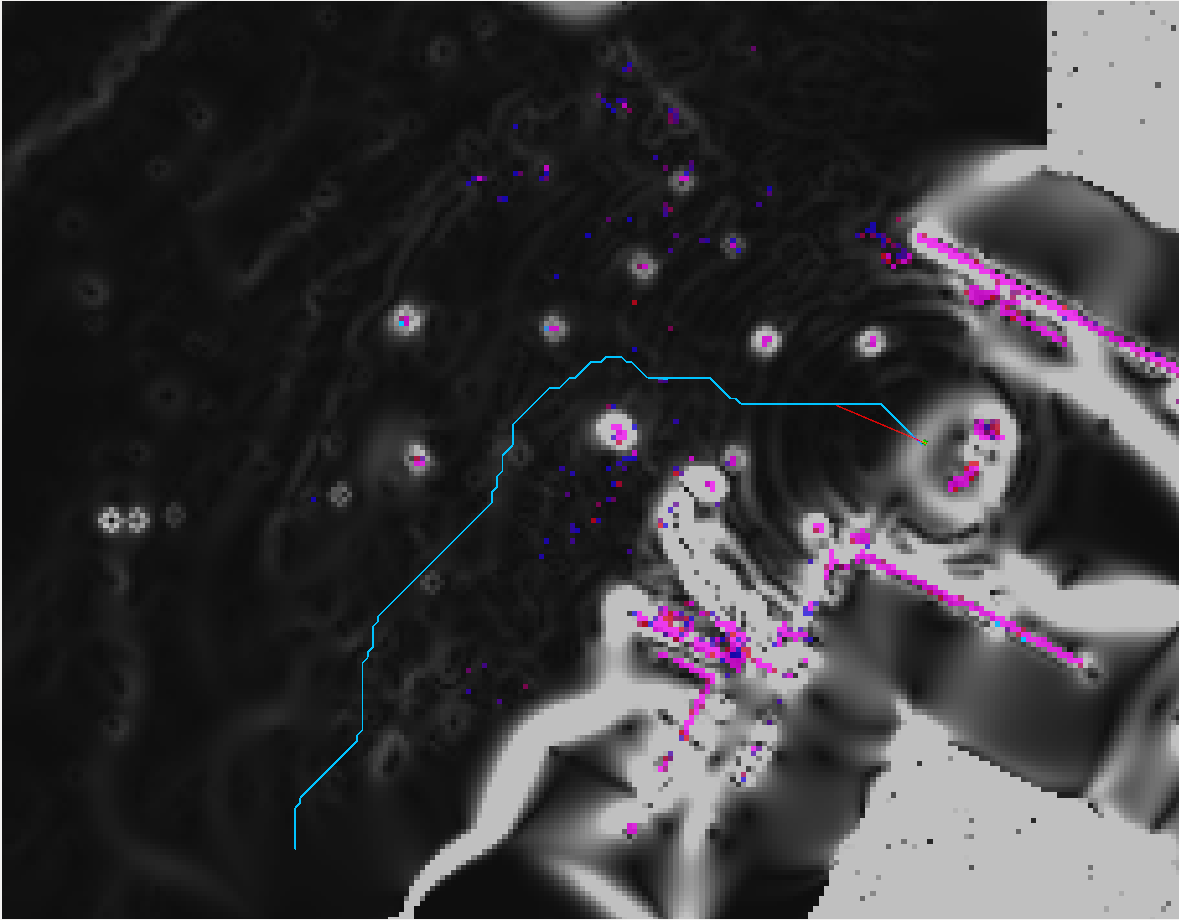}
      \includegraphics[width=\linewidth]{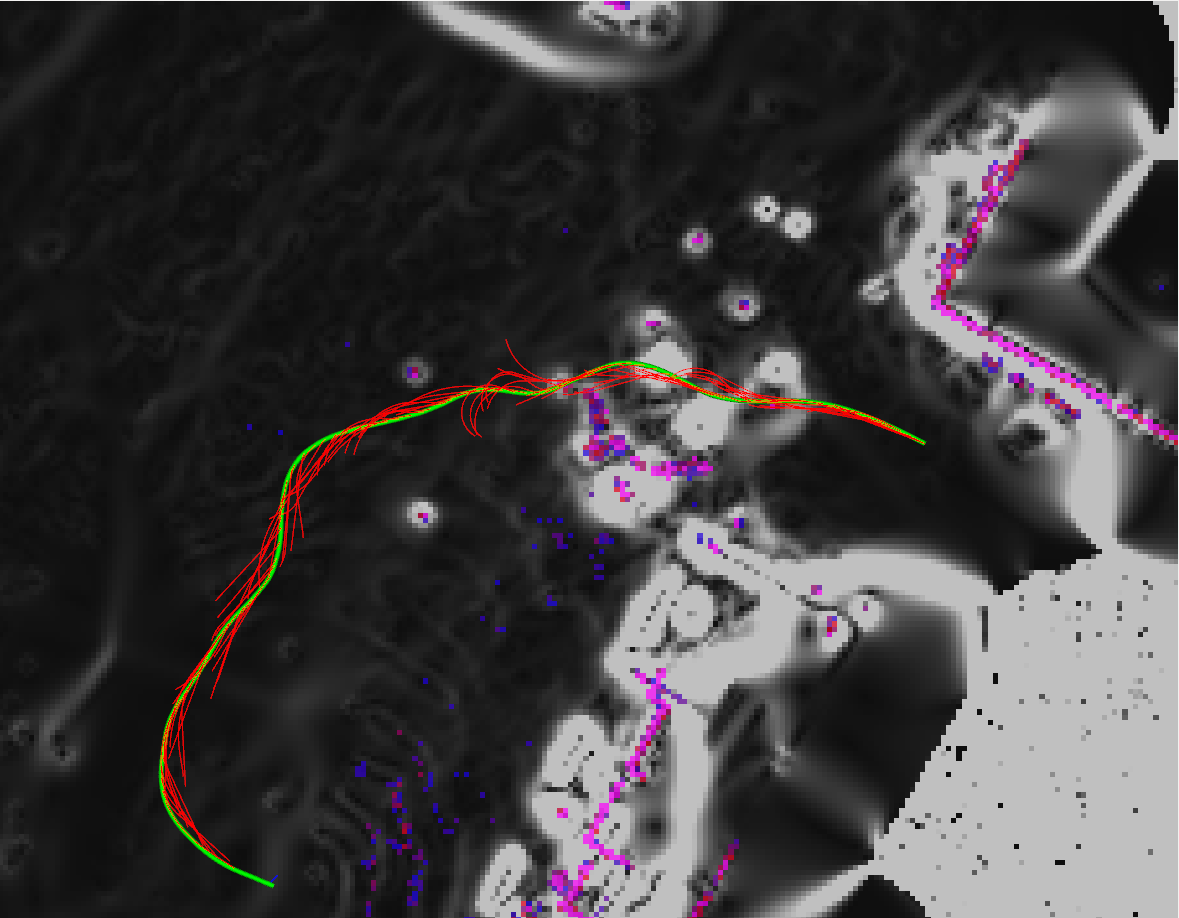}
	}}
      
      \caption{Robot driving through traffic cones placed in a field. Initial plan (top) and final path (bottom) A* path in teal, trajectories in red, and robot path in green}
      \label{test_1}
   \end{figure}
   
We implemented this algorithm on a Clearpath Robotics Warthog with a VLP-32c lidar and a Vectornav VN-300 IMU/GPS (figure~\ref{warthog}). The controller is an iterative linear quadratic regulator (ILQR) based controller described in~\cite{ilqr}. On our setup we were able to run the mapping module and A* planner at a rate of 5~Hz with the trajectory sampling module at 30~Hz. The Warthog has a maximum speed of 4.5~m/s. Testing was done at the Texas A\&M Rellis Campus. Here we present two tests done at 3~m/s, close to the maximum vehicle speed. The first, as shown in figure~\ref{test_1}, was done in a relatively flat field with several traffic cones as obstacles (figure~\ref{rellis}). The top of figure~\ref{test_1} shows the initial path in blue. The bottom shoes all the chosen trajectories in red and the final path in green. Note that the vehicle does not follow the initial path but rather finds a new path when it overshoots the original path. 

   \begin{figure}[thpb]
      \centering
      \framebox{\parbox{3in}{
      \includegraphics[width=\linewidth]{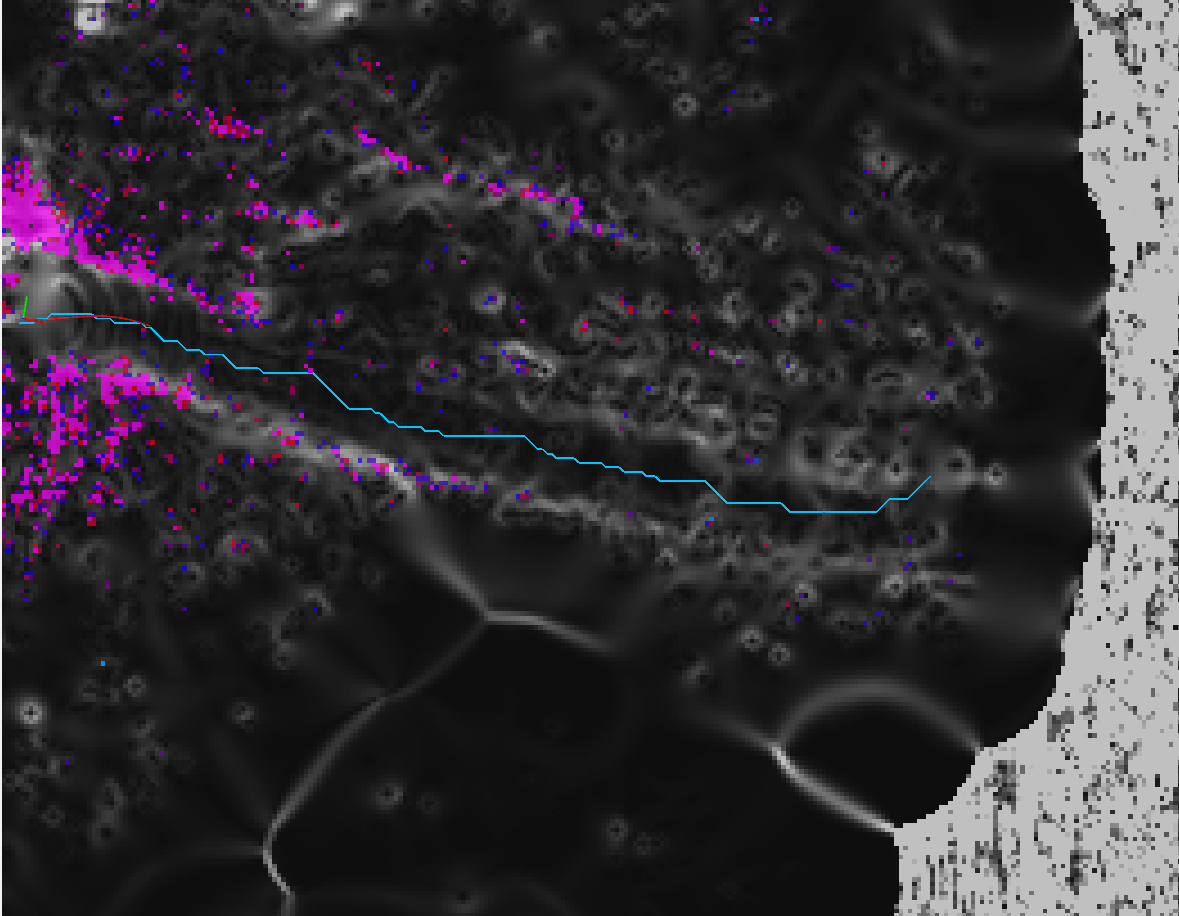}
      \includegraphics[width=\linewidth]{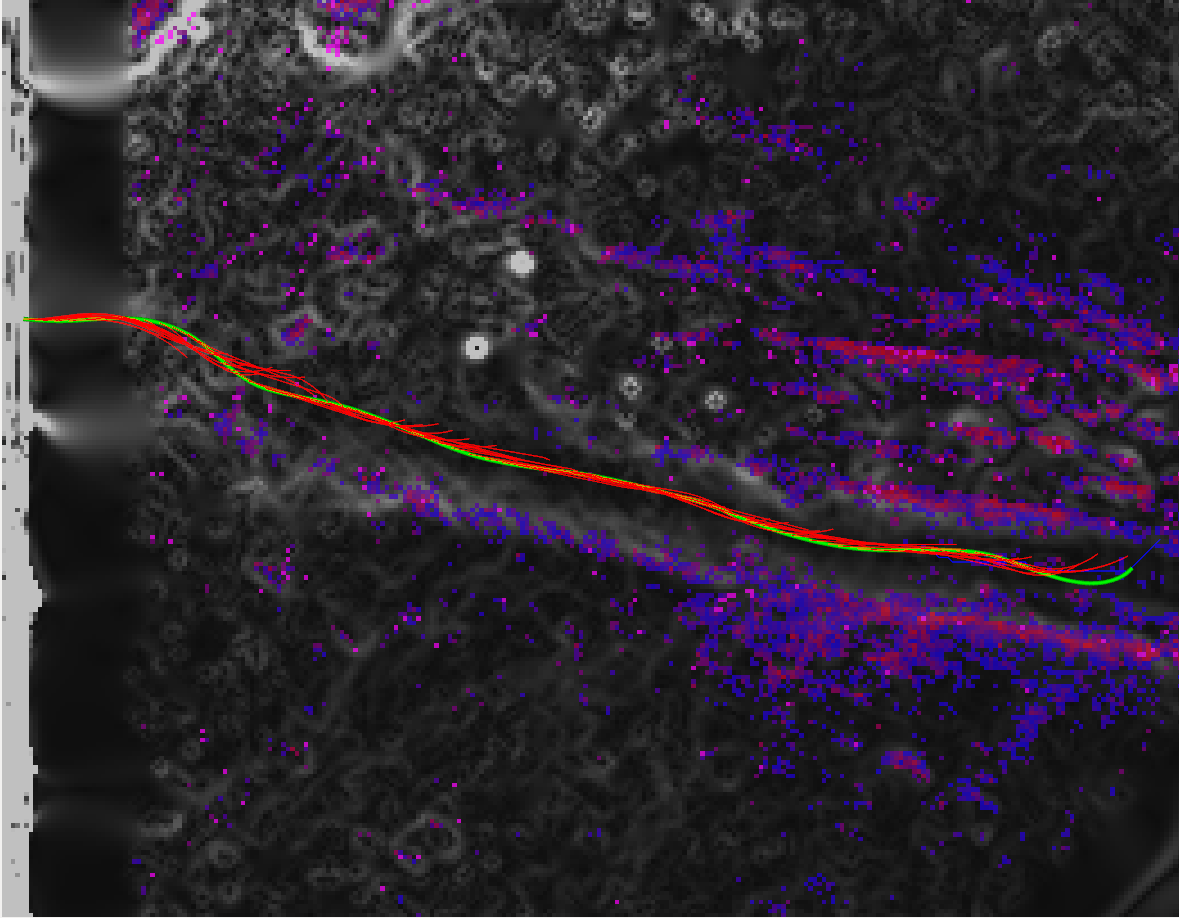}
	}}
      
      \caption{Robot driving along a approximately 50~m off-road path. Initial plan (top) and final path (bottom) A* path in teal, trajectories in red, and robot path in green}
      \label{test_2}
   \end{figure}
   
Figure~\ref{test_2} shows the second test presented here. It was an approximately 50~m path along an off-road trail with tall grass and brush along the edges. In this case the final path is very similar to the initial path. Minor deviations are made due to small changes in the local terrain gradient as more information is available.

\section{CONCLUSION}

We have presented a local planner that is able to make kinematically feasible paths across an unknown off-road environment. We combined the classic A* algorithm along with a kinematic forward trajectory simulation to generate good enough paths. Our experiments have shown this planner to work in real time at speeds of 3 m/s.
In the future we would like to integrate semantic segmentation so as to identify moving obstacles and plan around them. Additionally, this needs to be integrated with a global planner. Currently, if our planner is given a target point outside the ROI it will attempt to reach it but the produced global path will be greedy.
 
\balance

\bibliographystyle{IEEEtran}
\bibliography{IEEEabrv,IEEEexample}

\end{document}